\documentclass[a4paper]{article}
\usepackage{MM805}
\usepackage[fontsize=12pt]{fontsize}
\author{Zhipeng Chang(zchang) \\
Ruiling Ma(ruilin3) \\
Wenliang Jia(wenliang)}
\title{Pedestrain detection for low-light vision}

\begin{document}
\maketitle

\section{Short Abstract}
The demand for pedestrian detection has created a challenging problem for various visual tasks such as image fusion. As infrared images can capture thermal radiation information, image fusion between infrared and visible images could significantly improve target detection under environmental limitations. In our project, we would approach by pre-processing our dataset with image fusion technique, then using Vision Transformer (ViT) \cite{VIT} model to detect pedestrians from the fused images. During the evaluation procedure, a comparison would be made between YOLOv5 and the revised ViT model’s performance on our fused images.

\section{Brief summary of what exists}
Pedestrian detection for low-light vision has been extensively researched and developed. There are a lot of image fusion and object detection techniques that already exist, providing a wide range of options for researchers.

\subsection{Image fusion}
\subsubsection{Jia et al \cite{LLVIP} tested several image fusion techniques in their work, including GTF, FusionGAN, Densefuse, and IFCNN:}

\begin{enumerate}
\item Gradient Transfer Fusion is a method based on gradient transfer and total variation minimization. The method involves transferring the gradient information from the visible image to the infrared image, which is then fused with the original visible image using total variation minimization. This process helps to preserve the edges and fine details of the visible image while enhancing the overall contrast and texture of the fused image. The technique has been shown to produce high-quality results and outperform other state-of-the-art fusion methods in terms of both visual quality and quantitative metrics \cite{GTF}.

\item FusionGAN is an end-to-end image fusion technique that uses a generative adversarial network (GAN) to combine infrared and visible images, which has two networks: a generator network and a discriminator network. The generator network is responsible for generating a fused image that combines significant infrared intensities with additional visible gradients, while the discriminator network ensures that the fused image has more details from the visible image. Unlike traditional methods, FusionGAN does not require manually designing complex activity level measurements or fusion rules. This approach has been demonstrated to produce high-quality fused images with superior visual quality and improved detail preservation \cite{FusionGAN}.

\item DenseFuse utilizes a dense block architecture and multi-scale fusion to preserve fine details and enhance image quality. The approach consists of three parts: an encoder, a fusion layer, and a decoder. The input of the encoder is the source images, which are infrared and visible images. The encoder then utilizes CNN layer and dense block to obtain feature maps, which are fused using the addition and L1-norm fusion strategy. After the fusion layer, the fused feature maps are integrated into one feature map that contains salient features from both source images, and the fused image is reconstructed using the decoder network. Experimental results demonstrate that DenseFuse exhibits state-of-the-art fusion performance \cite{DenseFuse}.

\item IFCNN is a general image fusion framework based on convolutional neural network. IFCNN consists of three modules: feature extraction, feature fusion, and image reconstruction. The feature extraction module uses pre-trained CNN models to extract features from the source images. The feature fusion module combines the features using an appropriate fusion rule based on the input images. The image reconstruction module employs two
convolutional layers to generate the fused image. The proposed IFCNN framework outperforms other state-of-the-art methods in terms of both objective metrics and visual quality for various image fusion tasks \cite{IFCNN}.
\end{enumerate}

\subsubsection{Linfeng et al \cite{SeAFusion} have also proposed a new image fusion technique called SeAFusion:}

\begin{enumerate}
\item SeAFusion is a real-time infrared and visible image fusion framework using semantic awareness. In order to enhance the description ability of the fusion network for fine-grained details, a gradient residual dense block was employed, along with an elaborate content loss, it’s able to effectively maintain salient target intensity and preserve texture detail. Additionally, a semantic loss was introduced to allow high-level semantic information to flow back to the image fusion module, improving performance in high-level vision tasks. A joint low-level and high-level adaptive training strategy was also introduced in order to achieve impressive performance in both image fusion and various high-level vision tasks \cite{SeAFusion}.
\end{enumerate}

\begin{figure}[!htb]
\centerline{
  \includegraphics[width=0.9\textwidth]{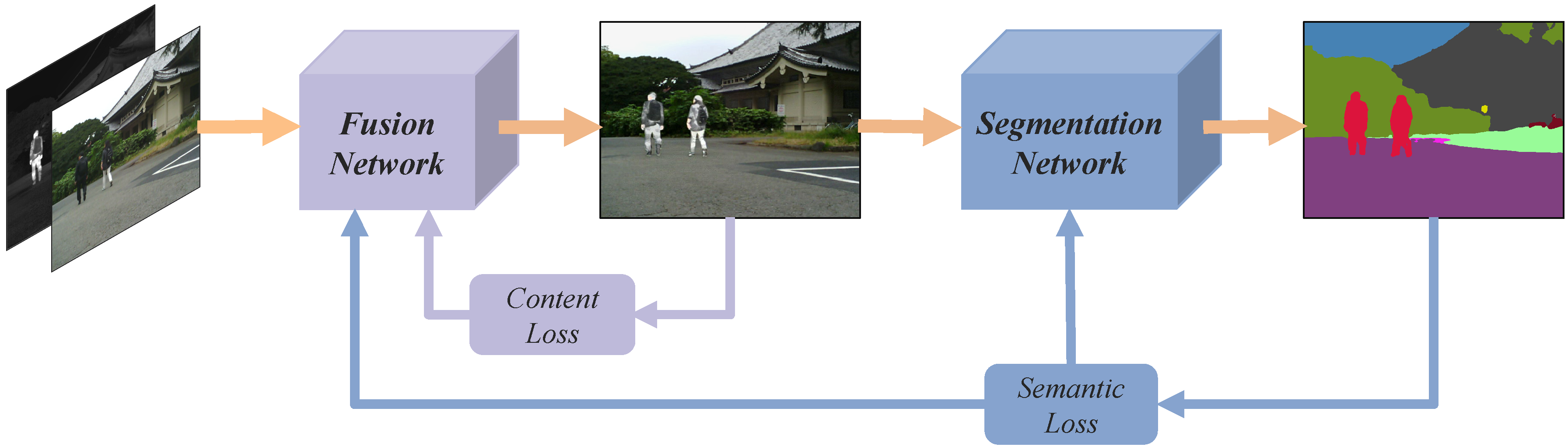}}
  \caption{The overall framework of the proposed semantic-aware infrared and visible image fusion algorithm \cite{SeAFusion}}
  \label{email_upload}
\end{figure}

\subsection{Object detection}
\subsubsection{Jia et al mainly tested YOLOv3 and YOLOv5: \cite{LLVIP}}

\begin{enumerate}
\item YOLO (You Only Look Once) is an object detection system that uses a single convolutional neural network to predict bounding boxes and class probabilities for objects in an image. YOLOv3 is an improved version of YOLO that uses darknet-53 as the backbone architecture, resulting in better performance than previous versions. YOLOv5 is a more recent version of YOLO that uses a new CSPDarknet53 with a focus on model size reduction and improved speed. YOLOv5 achieves better performance than YOLOv3 \cite{yolo}.
\end{enumerate}

\subsubsection{A new object detection technique called Vision Transformer (ViT) has been introduced and extensively studied in recent years}

\begin{enumerate}
\item ViT \cite{VIT} is an architecture based on the transformer architecture, originally developed for natural language processing tasks. According to the authors, this architecture is successful at image recognition tasks and achieves competitive results on standard benchmarks.
\end{enumerate}
\subsection{Dataset}
We will be using the LLVIP dataset presented by Jia et al \cite{LLVIP}. This dataset consists of 15,488 pairs of images that are precisely aligned in both time and space. It includes a significant number of pedestrians and most images are captured under low-light conditions.

\section{Project Plan identifying what aspects are novel}
Pedestrian detection from visible images, infrared images, and fused images is a well-developed process. However, there is currently no study that has utilized ViT as the main object detection technique. In our project, we aim to use ViT as the primary object detection technique to detect pedestrians from fused images. Therefore, the process of fusing images and then using ViT to detect pedestrians is novel. 

\section{Milestones}
There are two major milestones for this project:
\begin{enumerate}
\item {\bf Selecting an appropriate image fusion technique for the project, and fuse the images}
\item {\bf Using ViT to detect pedestrians from fused images}
\end{enumerate}	

\section{Breakdown of who will do what}
\begin{enumerate}
\item {\bf Setting up the necessary infrastructure and reproducing the existing work described in relevant papers to establish a baseline:} Zhipeng Chang
\item{\bf Selecting an appropriate image fusion technique and fuse testing images:} Wenliang Jia
\item {\bf Utilizing ViT technique to detect pedestrians:} Ruiling Ma
\end{enumerate}	

\section{Literature Review}
Medical image registration involves several techniques, including control-point-based registration, moment-based registration, contour-based registration, and optimization-based registration. Cheng et al. used the contour-based registration method to segment the upper airway, as it typically produces accurate results. To track the contour of the airway, they developed a method that uses modified GVF snakes with extra edge detection and snake shifting techniques, along with the contour of an adjacent CT slice to track the current slice. One key advantage of their method is that it is fully automated, making it more efficient than other methods. Medical experts evaluated their results for accuracy and found them to be acceptable, highlighting the success of their approach \cite{1}.

The cost of setting up per-patient based environments is one of the challenges in current rehabilitation methods. To overcome this issue, Rossol et al. developed a Virtual Reality solution using a stand-alone open-source 3D application available on multiple platforms. They also developed an algorithm based on Bayesian networks that can adapt the training simulation based on each patient's skill level. This intelligent adaptation of the training simulation reduces the high costs associated with traditional rehabilitation methods. The effectiveness of their solution was evaluated through small-scale user evaluations, which showed that their proposed application outperforms traditional power wheelchairs. However, the authors acknowledge that other factors should also be considered in their study \cite{2}.

Accurate localization and tracking of facial features are essential for developing high-quality model-based coding systems, such as MPEG-4. Bernogger et al. have introduced a new algorithm that significantly improves the accuracy of well-known facial feature detection and tracking algorithms. Additionally, they have presented animation models for eye movements to better visualize their results. The process involves synthesizing eye movement by computing the deformation of the 3D model's eyes using the extracted eye features. First, they create an individualized 3D face model and map it to the first frame of the face sequence. Then, they apply deformation parameters to the 3D model based on the extracted eye features to synthesize eye movement in successive frames. They evaluate the developed algorithms through experiments using various color images of different eyes and eye sequences. The results are promising for synthesizing eye movements, indicating the effectiveness of the proposed approach \cite{3}.

Due to the limitations of existing conventional imaging systems, capturing images or videos with a large dynamic range often leads to overexposed or underexposed results. Additionally, current approaches may face artifacts in their reconstructions when there is a high contrast or colored light source in the scene. To address these issues, the author proposes a novel neural network model that combines self-attention mechanisms, adversarial training, and a tailored loss function. These methods help the network take advantage of location interdependency and minimize artifacts. The author's image enhancement model outperformed other recent methods in terms of image naturalness and adaptability to various lighting conditions, as demonstrated through full-reference and non-reference image quality assessments \cite{4}.

Camera calibration is essential for the process of relating a 2-D image to a 3-D environment. Existing linear techniques are simpler to implement but cannot model camera distortions, while non-linear methods can consider complicated imaging models but require computationally expensive search procedures. In this paper, the authors present two alternative algorithms. The first method uses perspective distortions, but it is not reliable for real or synthetic data with low levels of noise. The second method is introduced to address this problem, and the article demonstrates that it produces reliable estimates for synthetic data with high levels of noise and real scenes. The proposed methods have achieved both mathematical simplicity and high accuracy results when synthetic data was used. By using these algorithms, an active camera can automatically calibrate itself \cite{5}.

The demand for imaging capabilities in various industries has led to the need to fuse infrared images with visible light. Although the benefits of infrared images and visible light imaging techniques can be combined through fusion, existing techniques struggle to maintain details. To address this issue, this paper proposes a multi-scale Gaussian rolling guidance filter decomposition-based method for fusing infrared and visible images. Initially, the source images are separated into three layers: the detail preservation layer, the edge preservation layer, and the energy base layer. Based on the unique traits of each layer, three fusion strategies are employed. The final fusion image is produced by combining the results of the three separate scales, resulting in additional valuable scene information. The author demonstrates that the proposed fusion framework outperforms most evaluations, based on subjective and objective assessments of multiple sets of fusion outcomes. However, the method's comparatively high complexity results in inferior real-time performance \cite{6}.

Motion capture data can effectively produce skeletal animation, and an efficient compression algorithm can guarantee the quality of the animation while transmitting the data. However, some coding techniques require high processing resources for compression or decompression and may not yield good results if the distortion metric is not dependent on the joint hierarchy. To address this issue, Firouzmanesh, Cheng, and Basu proposed an efficient encoding and decoding technique for lossy compression of motion capture data. This method takes human perception into consideration and focuses on bone lengths and variation in rotation. It guarantees that the motion data is preserved and transmitted in an equivalent or better perceptual quality under limited bandwidth \cite{7}.

As smart rooms continue to develop, efficient algorithms are required to achieve more precise human motion analysis and pose recognition. Currently, pose recognition can be generally separated into model-based systems and non-model based systems. To improve the accuracy of pose recognition, Singh, Mandal, and Basu introduced two pattern recognition methods, the Hough transform (HT) and the Radon transform (RT), into their algorithm. However, the use of HT can result in distortion in the shape and location of peaks in transform space, as well as problems while implementing the discrete HT via accumulator bins. Therefore, Singh, Mandal, and Basu proposed using the RT in their human poses and gestures recognition algorithm. The experimental results indicated that this algorithm guarantees a higher recognition rate for human arm and leg poses without normalizing the image in the pre-processing stage \cite{8}.

Intraspinal microstimulation (ISMS) involves inserting microwires from the back of the spinal cord to stimulate nerve cells and help people restore some leg functions after an injury. However, there are still some potential risks associated with using ISMS that cannot be ignored. To help clinicians minimize damage to the spinal cord system, computed tomography or MRI scans are used to identify an optimal path for microwire implantation. In the preprocessing stage, Mukherjee et al. proposed an automatic spinal cord image segmentation technique to assist with surgical planning. Compared to a semi-automatic spinal cord image segmentation technique, the proposed method uses active tracing of the segment boundary to provide a better-quality vision of both the region of interest and its neighbours \cite{9}.

Yin and Basu proposed a three-step method for implementing a feature detection method on facial organ areas. The first step involves limiting the facial feature detection region to certain areas, and a two-stage region-growing algorithm is used to make the detection less sensitive to noise. In the first stage, a large global growing threshold is used to explore more skin areas, while in the second stage, region information is estimated in the individual feature areas. The second step involves using pre-defined templates to extract the shape of the nostril and nose-side. Finally, after detecting the facial features and extracting the nose's shape, a 3D wireframe model is used to track the motion of facial expressions. The experimental results demonstrate that the proposed method can detect most nose features correctly \cite{10}.

The recognition of human action and group activity presents one of the biggest challenges in video understanding. To achieve a better comprehension of individual actions and group activity, Actor Relation Graph (ARG)-based group activity recognition is commonly employed. In order to improve the understanding of a group's activity in a single video scene, Kuang and Tie suggested using MobileNet in the CNN layer and Normalized Cross-Correlation and Sum of Absolute Differences to construct ARG. Moreover, they introduced a visualization model that uses bounding boxes on each human object to recognize and predict human action and group activity. Experimental results indicated that this method has the potential to improve the accuracy and speed of group activity recognition. However, Kuang and Tie acknowledged that further improvements are necessary to precisely identify individual actions \cite{11}.

To effectively display panoramic images in a telepresence system, Baldwin, Basu, and Zhang utilized a predictive Kalman filter. This system integrates a panoramic imaging system with a viewing direction predicted by the Kalman filter. The research by Baldwin, Basu, and Zhang presents experimental results from the prediction process and operator experience. They emphasized that improving the prediction process and obtaining better quantitative results require appropriate error modelling and parameter identification. When transmitting images from a remote location with delays, the system uses local image information to simulate a continuously flowing sequence of pictures \cite{12}.

Detecting straight lines using a pinhole camera model becomes difficult in cameras where the mirror system does not meet the Single Viewpoint (SVP) criteria. To solve this problem and detect straight-line features in panoramic non-SVP images, Fiala and Basu proposed an approach that uses the Panoramic Hough transform. The transform is specifically designed for catadioptric panoramic sensors with spherical mirrors. According to experimental results, this approach can identify horizontal lines to complement the trivial detection of vertical lines, and it can robustly detect straight-line features with estimated calibration \cite{13}.

Shen, Cheng, and Basu developed an algorithm for multi-exposure fusion using the hierarchical multivariate Gaussian conditional random field model to enhance its performance. The model integrates two perceptual quality metrics: perceived local contrast and color saturation. To preserve details, the model calculates contrast in the luminance channel of the LHS color space. Moreover, to ensure satisfactory results for color images, the model uses color saturation to measure the colorfulness of a pixel, representing the red, green, and blue components in the RGB color space. The experiment showed that the proposed algorithm significantly improves the quality of fused images, providing viewers with a better quality of experience \cite{14}.

Due to the projected increase of the senior population to two billion by 2050, there is a high demand for smart home applications that include fall detection. Several options for human fall detection exist, such as wearable devices, floor pressure sensors, or single RGB, multiple cameras, or depth cameras. Hajari and Cheng proposed a real-time automatic fall detection technique that uses computer vision to analyze motion deformation and velocity of the action, with a single RGB camera. This method extracts the human head and center of mass from each video frame and uses a robust ROI detection technique to detect areas of interest. To detect falls, a set of thresholds and the vector of angle size and angular velocity are used. Experimental results demonstrate that this method is reliable and is not affected by lighting changes or camera lens distortion \cite{15}.

\clearpage	
\bibliographystyle{plain}
\bibliography{MM805/mm805}
\end{document}